\documentclass[10pt,twocolumn,letterpaper]{article}
\usepackage{iccv}
\usepackage{times}
\usepackage{epsfig}
\usepackage{graphicx}
\usepackage{amsmath}
\usepackage{amssymb}
\usepackage{multirow}
\usepackage{color}

\usepackage[pagebackref=true,breaklinks=true,letterpaper=true,colorlinks,bookmarks=false]{hyperref}
\iccvfinalcopy 
\def\iccvPaperID{1} 

\ificcvfinal\fi
\begin{document}

\title{3D High-Fidelity Mask Face Presentation Attack Detection Challenge 
}
\author{
	Ajian Liu$^{\rm 1 }$, 
	Chenxu Zhao$^{\rm 2}$, 
	Zitong Yu$^{\rm 3}$, 
	Anyang Su$^{\rm 2}$, 
	Xing Liu$^{\rm 2}$, 
	Zijian Kong$^{\rm 2}$ \\
	Jun Wan$^{\rm 4,5,9}$\thanks{Corresponding author}, 
	Sergio Escalera$^{\rm 7,9}$, 
    Hugo Jair Escalante$^{\rm 8,9}$, 
    Zhen Lei$^{\rm 4,5,6}$, 
	Guodong Guo$^{\rm 10}$ \\ 
	$^{\rm 1}$MUST, Macau \quad 
	$^{\rm 2}$Mininglamp Academy of Sciences, Mininglamp Technology, China \\ 
	$^{\rm 3}$University of Oulu, Finland \quad 
	$^{\rm 4}$NLPR, CASIA, China \quad 
	$^{\rm 5}$SAI, UCAS, China \quad 
	$^{\rm 6}$CAIR, HKISI, CAS \quad \\ 
	$^{\rm 7}$CVC, UB, Spain 
	$^{\rm 8}$INAOE, CINVESTAV, Mexico \quad 
	$^{\rm 9}$ ChaLearn, USA \quad 
	$^{\rm 10}$Baidu Research, China \\ 
	\tt\footnotesize
	ajianliu92@gmail.com, 
	\tt\footnotesize
	zhaochenxu@mininglamp.com, 
	\tt\footnotesize
	zitong.yu@oulu.fi \\
	\tt\footnotesize
	jun.wan@ia.ac.cn, 
	\tt\footnotesize
	sergio@maia.ub.es
}

\maketitle
\ificcvfinal\thispagestyle{empty}\fi

\begin{abstract}
The threat of 3D masks to face recognition systems is increasingly serious and has been widely concerned by researchers. To facilitate the study of the algorithms, a large-scale High-Fidelity Mask dataset, namely CASIA-SURF HiFiMask (briefly HiFiMask) has been collected. Specifically, it consists of a total amount of $54,600$ videos which are recorded from $75$ subjects with $225$ realistic masks under $7$ new kinds of sensors~\cite{liu2021contrastive}. Based on this dataset and Protocol 3 which evaluates both the discrimination and generalization ability of the algorithm under the open set scenarios, we organized a 3D High-Fidelity Mask Face Presentation Attack Detection Challenge to boost the research of 3D mask-based attack detection. It attracted $195$ teams for the development phase with a total of $18$ teams qualifying for the final round. All the results were verified and re-run by the organizing team, and the results were used for the final ranking. This paper presents an overview of the challenge, including the introduction of the dataset used, the definition of the protocol, the calculation of the evaluation criteria, and the summary and publication of the competition results. Finally, we focus on introducing and analyzing the top ranking algorithms, the conclusion summary, and the research ideas for mask attack detection provided by this competition.
\end{abstract}

\section{Introduction}
Recently, Face Anti-Spoofing (FAS) has been attracted more and more attention~\cite{yu2021deep,yu2021revisiting,qin2021meta} due to the wide application of face recognition in financial payment, access control, and phone unlocking. Therefore, face Presentation Attack Detection (PAD) technology is a critical stage to reinforce the face recognition systems by determining whether the captured face from an imaging sensor is real or fake. With the release of several high-quality 2D attack datasets~\cite{Boulkenafet2017OULU,Liu2018Learning,zhang2019dataset,liu2021casia}, the previous algorithms~\cite{yu2020auto,yu2021dual,wang2020deep,yu2020searching,qin2019learning,yu2020face,zhang2021structure,liu2021dual,chen2021generalizable,zhang2020face,zhang2021aurora,liu2021adaptive} show better performance against Print attacks and video Replay attacks, but they are more vulnerable to mask attacks with more realistic color and structure. However, with the maturity of 3D printing technology, face mask has become a new type of Presentation Attack (PA), which can easily fool the FAS system based on coarse texture and facial depth information. 

Although some works have been devoted to 3D mask attacks, including datasets~\cite{nesli2013spoofing,liu20163d,li2018unsupervised,george2019biometric,yu2020nasfas} collection and algorithms~\cite{zhang2011face,kose2014mask,steiner2016reliable,liu20163d,li2016generalized,liu2018remote,lin2019face} design, there are still some limitations that hinder the performance of the algorithm: (1) Lack of a high-quality and large-scale mask dataset for algorithm research, due to the limitation of high fidelity mask production cost. As far as we know, the existing mask datasets are insufficient in the aspects of the number of subjects and skin tones, mask quality and types, scene settings, lighting environments, and collection devices, which seriously limits the research of data-driven algorithms. (2) Lack of a challenging and public benchmark for performance comparison of different algorithms. As a result, the existing algorithms only work for specific mask types or in constrained environments. Several rPPG-based methods~\cite{li2016generalized,liu20163d,liu2018remote,lin2019face,liu2021multi,yu2021transrppg} are proposed according to the evidence that periodic rPPG pulse cues could be recovered from the live faces but noisy for the mask attacks. However, they are vulnerable to the interference of illumination change and sensitive to detection distance. (3) Compared with 2D attacks, such as Print-Attack, Replay-Attack, high fidelity mask has realistic skin color and structure. It is difficult to distinguish between a live face and a mask from the visible spectrum.

In order to promote the community's research on mask attack detection, we solve the current difficulties from the following three aspects based on the above analysis: (1) We collect and release a large-scale 3D high-fidelity mask face PAD dataset named HiFiMask. Compared with public 3D mask datasets, it has several advantages, such as high fidelity masks and amount of data in the term of identities, lightings, sensors, and videos. (2) We define a more general and valuable testing protocol for real-world deployment and provide a decent result as a benchmark. Our protocol evaluates both the discrimination and generalization ability of the algorithm under the open set scenarios. In other words, the training and developing sets contain only parts of common mask types and scenarios while there are more general mask types and scenarios on the testing set. (3) Based on the dataset and protocol, we successfully held a competition, \emph{3D High-Fidelity Mask Face Presentation Attack Detection Challenge at ICCV2021}~\footnote{\url{https://competitions.codalab.org/competitions/30910}}, attracting $195$ teams from all over the world. The results of the top three teams are far better than our baseline results, which greatly pushes the current best performance of mask attack detection. A summary with the names and affiliations of teams that entered the final stage is shown in Tab.~\ref{table:team-affiliations}. Interestingly, compared with the previous challenges~\cite{2019Multi,liu2020cross,boulkenafet2017competition}, the majority of the final participants of this competition come from the industrial community, which indicates the increased importance of the topic for daily life applications.
\begin{table}[]
\footnotesize
\centering
\caption{Team and affiliations name are listed in the final ranking of this challenge.}
\begin{tabular}{|c|c|c|}
\hline
Ranking & Team Name                                                    & Leader Name, Affiliation                                                                                      \\ \hline \hline
1       & VisionLabs                                                   & Oleg Grinchuk, visionlabs.ai                                                                                  \\ \hline
2       & WeOnlyLookOnce                                               & Ke-Yue Zhang, Tencent Youtu Lab                                                                               \\ \hline
3       & CLFM                                                         & Samuel Huang, FaceMe                                                                                          \\ \hline
4       & oldiron666                                                   & \begin{tabular}[c]{@{}c@{}}Zezheng Wang, \\ Kuaishou Technology\end{tabular}                                  \\ \hline
5       & \begin{tabular}[c]{@{}c@{}}Reconova-\\ AI-LAB\end{tabular}   & \begin{tabular}[c]{@{}c@{}}Mingmu Chen, \\ Reconova Technology\end{tabular}                                   \\ \hline
6       & inspire                                                      & Jiang Hao, Bytedance Ltd.                                                                                     \\ \hline
7       & Piercing Eyes                                                & \begin{tabular}[c]{@{}c@{}}Hyokong, \\ National University of Singapore\end{tabular}                          \\ \hline
8       & msxf\_cvas                                                   & \begin{tabular}[c]{@{}c@{}}Liang Gao, MaShang \\ Consumer Finance Co.,Ltd\end{tabular}                        \\ \hline
9       & VIC\_FACE                                                    & Cheng Zhen, Meituan                                                                                           \\ \hline
10      & \begin{tabular}[c]{@{}c@{}}DXM-DI-\\ AI-CV-TEAM\end{tabular} & Weitai Hu, Du Xiaoman Financial                                                                               \\ \hline
11      & fscr                                                         & \begin{tabular}[c]{@{}c@{}}Artem Petrov, Peter the Great St.\\ Petersburg Polytechnic University\end{tabular} \\ \hline
12      & VIPAI                                                        & Yao Xiao, Zhejiang University                                                                                 \\ \hline
13      & reconova-ZJU                                                 & Zhishan Li, Zhejiang University                                                                               \\ \hline
14      & sama\_cmb                                                    & \begin{tabular}[c]{@{}c@{}}Yifan Chen, \\ Chinese Merchants Bank(CMB)\end{tabular}                            \\ \hline
15      & Super                                                        & \begin{tabular}[c]{@{}c@{}}Yu He, \\ Technische Universität München, \\ mytum\end{tabular}                    \\ \hline
16      & ReadFace                                                     & Zhijun Tong, ReadFace                                                                                         \\ \hline
17      & LsyL6                                                        & Dongxiao Li, Zhejiang University                                                                              \\ \hline
18      & HighC                                                        & \begin{tabular}[c]{@{}c@{}}Minzhe Huang, \\ Akuvox (Xiamen) Networks \\ Co., Ltd.\end{tabular}                 \\ \hline
\end{tabular}
\label{table:team-affiliations}
\end{table}

To sum up, the contributions of this paper are summarized as follows: (1) We describe the design of the \emph{3D High-Fidelity Mask Face Presentation Attack Detection Challenge at ICCV2021} challenge. (2) We organize this challenge around the HiFiMask datsaset, proving the suitability of such a resource for boosting research on the topic. (3) We report and analyze the solutions developed by participants. (4) We conclude the effective scheme of mask attack detection from the top-ranked algorithms and point out the research direction through this competition.

\section{Challenge Overview}
In this section, we review the organized challenge, including a brief introduction of the HiFiMask dataset, the challenge process and timeline, the challenge protocol, and evaluation metrics.

\noindent \textbf{HiFiMask Dataset.}\label{section:HiFiMask}
HiFiMask~\cite{liu2021contrastive} is currently the largest 3D face mask PAD dataset, which contains $54,600$ videos captured from $75$ subjects of three skin tones, including $25$ subjects in yellow, white, and black, respectively. For mask types, it contains $3$ high-fidelity masks for each identity, which are made of transparent, plaster, and resin materials, respectively. During the acquisition process, it considers $6$ complex scenes for video recording, \ie, White Light, Green Light, Periodic Three-color Light, Outdoor Sunshine, Outdoor Shadow, and Motion Blur. For each scene, there are $6$ videos under different lighting directions (\ie, NormalLight, DimLight, BrightLight, BackLight, SideLight, and TopLight) to explore the impact of directional lighting. Among them, there is periodic lighting within [0.7, 4]Hz for the first three scenarios to mimic the human heartbeat pulse, thus might interfere with the rPPG-based mask detection technology~\cite{li2016generalized}. Finally, $7$ mainstream imaging devices (\ie, iPhone11, iPhone X, MI10, P40, S20, Vivo, and HJIM) are utilized for video recording to ensure high resolution and imaging quality.

In order to facilitate the participating teams to use the dataset, we have carried out some data preprocessing steps. We remove irrelevant background areas from original videos, such as the part below the neck. After face detection, we sample 10 frames at equal intervals from each video. Finally, we name the folder of this video according to the following rule: $Skin\_Subject\_Type\_Scene\_Light\_Sensor$.

\noindent \textbf{Challenge Protocol and Data Statistics.}
In order to increase the challenge of the competition and meet the actual deployment requirements, we consider a protocol that can comprehensively evaluate the performance of algorithm discrimination and generalization. In other words, the training and developing sets contain only parts of common mask types and scenarios while there are more general mask types and scenarios on the testing set. Based on Protocol 1~\cite{liu2021contrastive}, we define training and development sets with parts of representative samples while a full testing set is used. Thus, the distribution of testing sets is more complicated than the training and development sets in terms of mask types, scenes, lighting, and imaging devices. Different from Protocol 2~\cite{liu2021contrastive} with only ‘unseen’ mask types, the challenge protocol considers both `seen' and ‘unseen’ domains as well as mask types, which are more general and valuable for real-world deployment.
\begin{table}[]
\centering
\caption{Statistical information for Challenge Protocol. `\#' means the number of videos. Note that 1, 2 and 3 in the third column mean Transparent, Plaster and Resin mask, respectively. Other columns refer to~\ref{section:HiFiMask} in a similar way.}
\scalebox{0.70}{
\begin{tabular}{|c|c|c|c|c|c|c|l|l|c|l|l|c|l|l|}
\hline
Subset & subj. & Mask    & Scene   & Light   & Sensor  & \multicolumn{3}{c|}{\# live} & \multicolumn{3}{c|}{\# mask} & \multicolumn{3}{c|}{\# all} \\ \hline
Train  & 45      & 1,3      & 1,4,6    & 1,3,4,6  & 1,2,3,4  & \multicolumn{3}{c|}{1,610}   & \multicolumn{3}{c|}{2,105}   & \multicolumn{3}{c|}{3,715}  \\ \hline
Dev    & 6       & 1,3      & 1,4,6    & 1,3,4,6  & 1,2,3,4  & \multicolumn{3}{c|}{210}     & \multicolumn{3}{c|}{320}     & \multicolumn{3}{c|}{536}    \\ \hline
Test   & 24      & 1$\sim$3 & 1$\sim$6 & 1$\sim$6 & 1$\sim$7 & \multicolumn{3}{c|}{4,335}   & \multicolumn{3}{c|}{13,027}  & \multicolumn{3}{c|}{17,362} \\ \hline
\end{tabular}
}
\label{tab:protocol-3}
\end{table}

In the challenge protocol, as shown in Tab.~\ref{tab:protocol-3}, all skin tones, part of mask types, such as transparent and resin materials (short for 1, 3), part of scenes, such as White Light, Outdoor Sunshine, and Motion Blur (short for 1, 4, 6), part of lightings, such as NormalLight, BrightLight, BackLight, and TopLight (short for 1, 3, 4, 6), and part of imaging devices, such as iPhone11, iPhone X, MI10, P40 (short for 1, 2, 3, 4) are presented in the training and development subsets. While all skin tones, mask types, scenes, lightings, and imaging devices are presented in the testing subset. For clarity, the dataset partition and video quantity of each subset of the challenge protocol are shown in Tab.~\ref{tab:protocol-3}.

\noindent \textbf{Challenge Process and Timeline.}
The challenge was run in the CodaLab\footnote{\url{https://competitions.codalab.org/competitions/30910}} platform, and comprised two stages as follows:

\textbf{Development Phase:}~(\emph{Started: April. 19, 2021 - Ended: in June 10, 2021}). During this phase, participants had access to labeled training data and unlabeled development data. Participants could use training data to train their models, and they could submit predictions on the development data. Training data was made available with samples labeled with the genuine, 2 types of the mask (short for 1, 3), 3 types of scenes (short for 1, 4, 6), 4 kinds of lightings (short for 1, 2, 4, 6) and 4 imaging sensors (short for 1, 2, 3, 4). Although the development data maintains the same data type as the training data, the label is not provided to the participants. Instead, participants could submit predictions on the development data and receive immediate feedback via the leader board.

\textbf{Final phase:}~(\emph{Started: June 10, 2021 - Ended: June 20, 2021}). During this phase, labels for the development set were made available to participants, so that they can have more labeled data for training their models. The unlabeled testing set was also released, participants had to make predictions for the testing data and upload their solutions to the challenge platform. The test set was formed by examples labeled with the genuine, and all skin tones, mask types (short for 1$\sim$3), scenes (short for 1$\sim$6), lightings (short for 1$\sim$6), and imaging devices (short for 1$\sim$7). Participants had the opportunity to make 3 submissions for the final phase, this was done with the goal of assessing the stability of their methods. Note that the CodaLab platform defaults to the result of the last submission.

The final ranking of participants was obtained from the performance of submissions in the testing sets. To be eligible for prizes, winners had to publicly release their code under a license of their choice and provide a fact sheet describing their solution.

\noindent \textbf{Evaluation Metrics.}
In this challenge, we selected the recently standardized ISO/IEC 30107-3\footnote{\url{https://www.iso.org/obp/ui/iso}} metrics: Attack Presentation Classification Error Rate (APCER), Normal Presentation Classification Error Rate (NPCER) and Average Classification Error Rate (ACER) as the evaluation metrics. The ACER on the testing set is determined by the Equal Error Rate (EER) thresholds on the development set. Finally, The value ACER was the leading evaluation measure for this challenge, and Area Under Curve (AUC) was used as additional evaluation criteria.

\section{Description of solutions}
\begin{figure}
\centering
\includegraphics[width=1.05\linewidth]{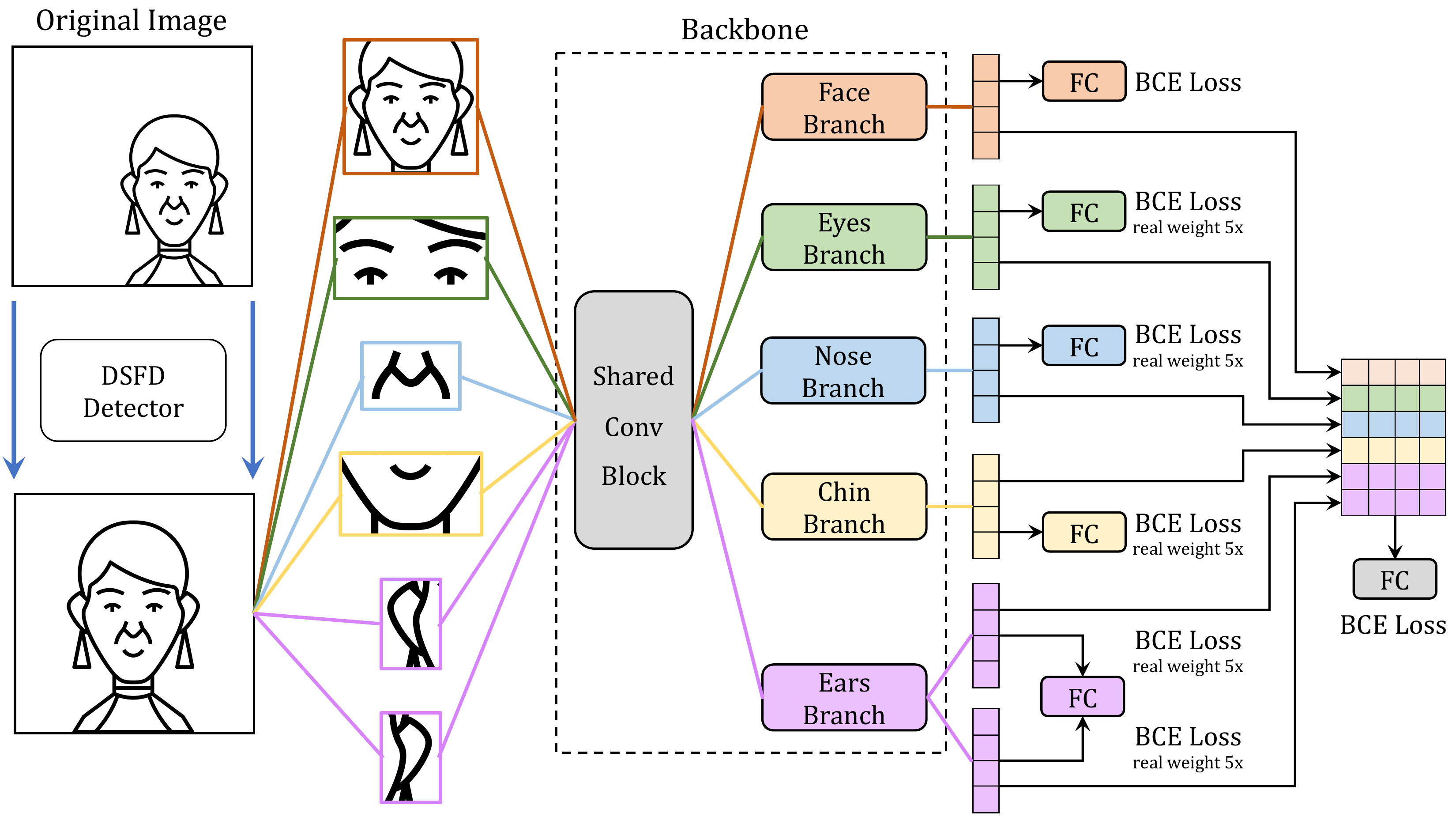}
\caption{\small{The pipeline of team VisionLabs. Original images are cropped by DSFD detector~\cite{DSFD} and split into five regions by prior knowledge. Then, those parts are input into the backbone with one shared convolution block and five branches. Each branch will output a 320-dimensional vector (two vectors from the ears branch). All vectors are concatenated to one vector and calculated as $L_{total}$.}}
\label{fig:VisionLabs}
\end{figure}

\noindent \textbf{VisionLabs}  \quad
Due to the tiny fake features of 3D face masks and the complexity to distinguish, team VisionLabs proposed a pipeline based on high-resolution face parts cropped from the original image, as shown in Fig.~\ref{fig:VisionLabs}. Those parts are used as additional information to classify the full images through the network.

During preparation state, centered face crops are created using the Dual Shot Face Detector (DSFD) detector~\cite{DSFD}. The crop bounding box is expanded $1.3\times$ times around face detection bounding box. If the bounding box is out of the original image border, missing parts are filled with black. If no face is found, the original image is used instead of crop. Then, five face regions are cropped using prior information from face bounding box, as eyes, nose, chin, left, and right ear. Each part will be resized to $224\times 224$ and input into the backbone after data augmentation (\ie, rotation, random crop, color jitter). Additionally, as a regularization technique, they turned 10\% of images into trash images by scaling random tiny parts of images. As shown in Fig.~\ref{fig:VisionLabs}, team VisionLabs used a multi-branch network, including Face, Eyes, Nose, Chin, and Ears branches. Since any pre-trained weight is prohibited in this competition, they tried to replicate the generalization ability of convolutional filters by using shared the first block of each branch. This made the first block filters learn more diverse features. Five branches all adopt EfficientNet-B0~\cite{efficient} as the backbone, and the original descriptor size of each branch is reduced from 1280 to 320. Due to the presence of left and right ears, the Ears Branch outputs two vectors. Then, each loss and confidence of the current branch can be obtained through the fully connected layer, as $L_{face}$, $L_{eyes}$, $L_{nose}$, $L_{chin}$, $L_{Lear}$,$L_{Rear}$ ($Lear$ for left ear and $Rear$ for right ear). These six vectors are concatenated to obtain a 1920-dimensional vector, used to calculate the loss function $L_{total}$. All branches are trained simultaneously with the final loss:
\begin{equation}
\begin{aligned}
    L= 5*L_{total}+5*L_{face}+L_{eyes}+L_{nose} \\
        +L_{chin}+0.5*L_{Lear}+0.5*L_{Rear}
     \label{con:VisionLab-loss}
\end{aligned}
\end{equation}
where all losses are binary cross-entropy(BCE) losses. Since face parts do not always contain the tiny fake features, for eyes, nose, chin and ears, they increase positive class weight in BCE loss by a factor of 5. The partial face part descriptors will not be punished too hard if don’t contain useful features. 

They trained the model with Adam optimizer for 60 epochs using an initial learning rate of 0.0006 and decreasing it every 3 epochs by a factor of 0.9. During the inference phase, they chose 0.7 as the test set threshold when it is inaccurate to select a threshold from a validation set close to the full score. Based on average positions and prior information, some face parts of images may be cropped in the wrong way. So a test time augmentation is introduced. They flip an image and obtain the final results by averaging the scores of original and flipped faces.

\vspace{0.5em}
\noindent \textbf{WeOnlyLookOnce}   \quad
In this method, considering that there are irrelevant noises in the raw training data, a custom algorithm is used to detect black borders firstly. After that, the DSFD~\cite{DSFD} is applied to detect potential faces for each image. To be specified, the training set is processed by wiping black borders merely, while the testing set and validation set are cropped with a ratio of 1.5 times bounding box further. What's more, the positive samples in the training set are much less than negative samples. The training augmentations include rotation, image crop, color jitter, etc.
\begin{figure}
\centering
\includegraphics[width=1.0\linewidth]{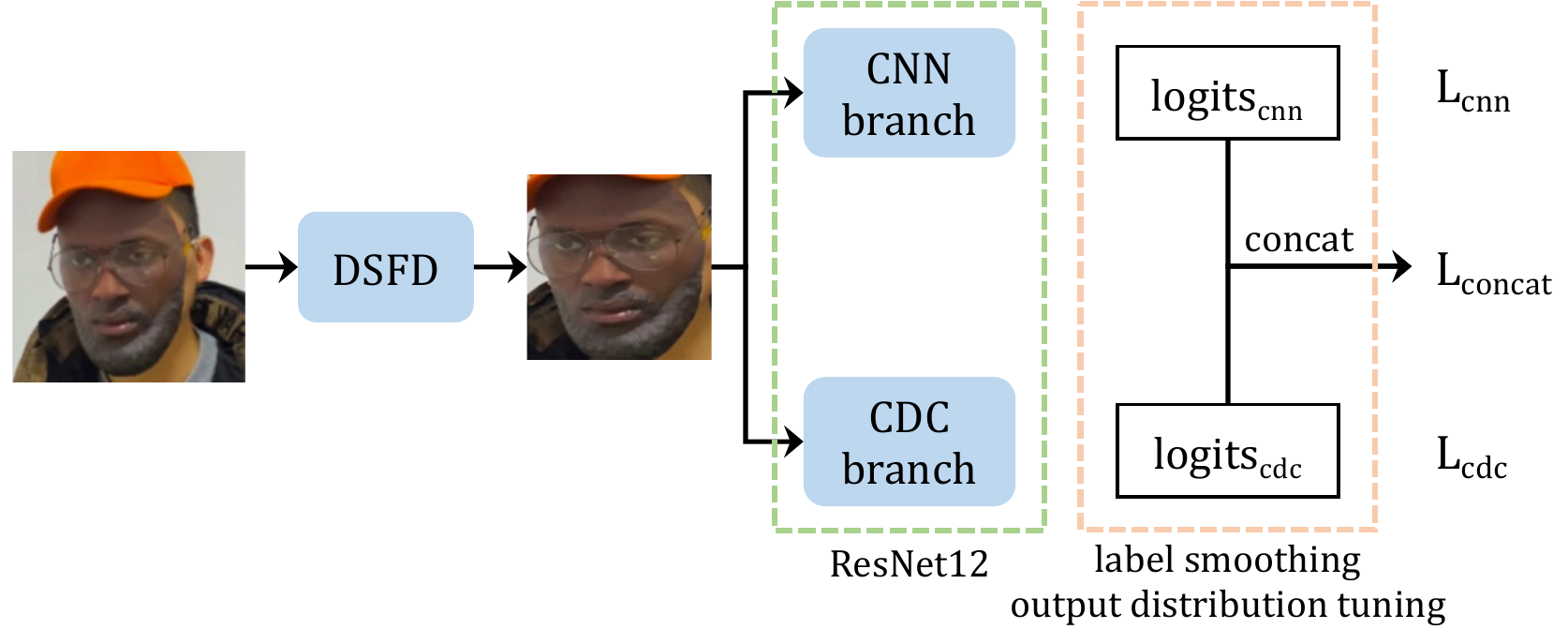}
\caption{\small{The framework of team WeOnlyLookOnce. The DSFD face detector is used to detect bounding box. Afterwards, a lightweight self-defined ResNet12 subsequently aims to classify the input into three categories. To be mentioned, Label smoothing and output distribution tuning are used as additional tricks.}}
\label{fig:tencent}
\end{figure}

As shown in Fig.~\ref{fig:tencent}, the framework \cite{wang2020face,chen2021local,wang2021delving,yin2021adv} consists of a CNN branch and a CDC branch. Both networks are self-designed lightweight ResNet12, and each of them is a three-class classification network aiming to detect real images and two kinds of mask. The realization of CNN branch is a vanilla convolution while the CDC branch used Central Difference Convolution \cite{yu2020searching}. To alleviate the overfitting problem, the team additionally adopted a label smoothing strategy and an output distribution tuning strategy inspired by temperature scaling \cite{NEURIPS2019_f1748d6b}. After computing the cross-entropy loss by using the logits and label, the total loss is calculated by the following equation
\begin{equation}
\begin{aligned}
    L= L_{concat} + 0.5 * L_{cnn} + 0.5 * L_{cdc}
    \label{con:tencent-loss}
\end{aligned}
\end{equation}

To minimize the distribution gap between validation and test sets, this team proposed an effective distribution tuner. They provide two strategies in this tuner both of which are proved to be effective. In the first strategy, they reform the three-class classification task into a binary classification task by adding the two attack-class logits into one uniformed value, then dividing the real logits by a factor of 3.6 and the fake logits by a factor of 5.0 before the softmax operation. In the second strategy, the task still remains a three-class classification problem while the real score on the validation set is subtracted by 0.07.

\vspace{0.5em}
\noindent \textbf{CLFM}  \quad
Team CLFM produced a model with only cross-entropy loss based on CDCN++ model but earn a good result. The central difference convolution was used to replace traditional convolution. Also, attention modules were introduced in each stage to make the model performed better. Besides, they fuse three stages' output parameters as feature vectors before the fully connected layer.

For data pre-processing, they adopt their own face detection model to perform face detection and take patches of the face as input. Something should be the highlight that they play some brilliant and practical tricks both on train and test set. On the one hand, they find that there are hats/glasses that will most likely lead the model in the wrong direction. So they firstly crop the face according to the bounding box and then crop the region around the mouth. The face size is randomly set in a small range to ensure the generalization of the model. If the region is not enough to fill it, they tend to flip and mirror the region to keep the texture constant. The model input is square blocks resized to $56\times 56$ and normalized with mean and standard deviation parameters which are summarized from ImageNet. On the other hand, they also notice that there are about 17\% of the images in the test set that the face detection didn't detect any face, and in this kind of scenario the model will have no other choice but use the whole image as the bounding box. So they randomly make part of the training data’s bounding box to be the whole image and make slight changes to the cropping process of the test set compared with the training set, so that at least the model won’t be pure guessing when facing this situation. Finally, they use self voting by moving the patch in a small range and averaging them as the final score.

\vspace{0.5em}
\noindent \textbf{Oldiron666} \quad
The team Oldiron666 proposed a self-dense regularization framework for face anti-spoofing. For data pre‐processing, they expand an adaptive scale for the cropped face, which improves the performance. The input size of the images is 256, and the following data augmentations are performed to improve generalization, such as Random Crop, Cutout, and Patch Shuffle Augmentation, etc.


The team Oldiron666 used a representation learning framework similar to SimSiam \cite{simsiam}, but introduces a multilayer perceptron(MLP) for supervised classification. During the training process, the face image \textsl{$X$} will be randomly augmented to obtain two views for input, $X_1$ and $X_2$. The two views are processed by an encoder network \textsl{$f$}, which consists of the backbone and an MLP head called projector~\cite{simCLR}. They found a more light network may bring better performance on the HiFiMask. Therefore, Resnet6 is utilized as the backbone, which contains a low computational complexity. The output of one view is transformed to match the other view by a dense predictor MLP head, denoted as $h$. The dense similarity loss, marked as $L_{contra}$, maximizes the similarity between both sides. To implement supervised learning, they perform the dense classifier $c$ at the end of the framework and use Mean Squared Error(MSE) to evaluate the output. MSE loss is calculated with the ground truth label on one side, denoted as $L_{cls}$, while $L_{d}$ is calculated as the difference between the category output of both two sides. The training loss can be defined as,
\begin{equation}
\begin{aligned}
    L= L_{contra}+L_{cls}+0.1*L_{d}
    \label{con:Oldiron666-loss}
\end{aligned}
\end{equation}
During the training process, they perform half-precision floating-point to obtain a faster training speed. The SGD optimizer is adopted, with an initial learning rate of 0.03, weight decay by 0.0005, and momentum by 0.9. During inference, only the $X_1$ side is executed to obtain the result of face anti-spoofing.

\begin{figure}
\centering
\includegraphics[width=1.0\linewidth]{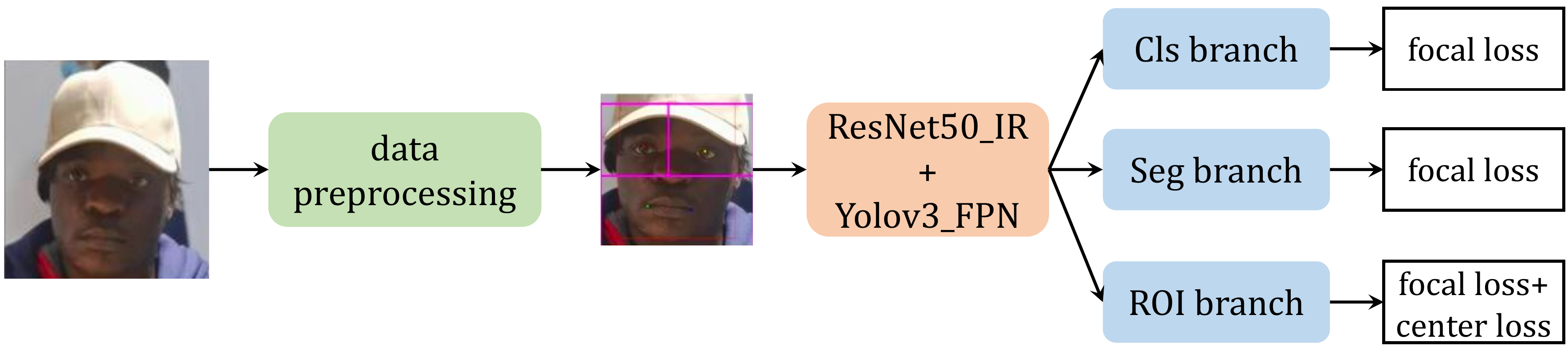}
\caption{\small{The application flow chart of team Reconova-AI-Lab. Raw images are firstly pre-processed by RetinaFace for face detection, crop and alignment in the upper flow. Then, they used a multi-task learning algorithm, which mainly includes three branches.}}
\label{fig:reconova}
\end{figure}

\vspace{0.5em}
\noindent \textbf{Reconova-AI-Lab} \quad
Team Reconova-AI-LAB contributes a variety of models and generates many different results, the best of which is used for the competition. They proposed a multi-task learning algorithm, which mainly includes three branches, the direct classification branch, the real person learning Gaussian mask branch, and the Region of interest(ROI) classification branch. In the rest of this section, we take Cls, Seg, and ROI branches as abbreviations respectively. Cls branch takes a focal loss which combining Sigmoid and BCE Loss as the supervision information. It is annotated as $loss\_classi$. Seg branch adopts the same loss function as Cls and its loss annotation is $loss\_seg$. Concerning with ROI branch, it take three loss functions, which is $loss\_cls1$, $loss\_cls2$ and $loss\_center$, respectively. The effect of the first one is focal loss mentioned before. The second one aims to the alignment of ROI which is used to calibrate the operation of ROI pooling. Subsequently, the purpose of the last one is to reduce the distance between classes. Finally, the lost function of ROI branch $loss\_roi$ equals $loss\_cls1$ plus $loss\_center*$ $\times 0.01$ plus $loss\_cls2$. All branches are trained synchronously with an SGD optimizer in 800 epochs, and the total loss function formulates as follows:
\begin{equation}
\begin{aligned}
    total\_loss = loss\_classi + loss\_seg + loss\_roi
     \label{con:Reconova-AI-LAB-loss}
\end{aligned}
\end{equation}

Their application flow chart is shown in Fig.~\ref{fig:reconova}. First, the data pre-processing includes the use of RetinaFace to detect the face and generate 14 landmarks per face, including face coordinates and bounding boxes of left, right ear, and mouth. At that stage, they use some strategies to avoid large-angle posture and non-existence of face by constraining the size of the bounding box of ROI. Meanwhile, they take mirroring, random rotation, random color enhancement, random translation, and random scaling as treatments of data enhancement. Then they adopt a backbone called Res50\_IR, which has stacked 3, 4, 14, and 3 blocks respectively in four stages. In order to enhance features, an improved Residual Bottleneck structure named Yolov3\_FPN is connected to the different stages of the network. The slightly complicated network is followed by three branches mentioned before. All of the parameters are initialized by different methods according to different layers.

\noindent \textbf{inspire}
The team firstly utilized a ResNet50 \cite{he2016deep} based RetinaFace \cite{deng2019retinaface} to detect face bounding boxes for all images. To be noticed, three different threshold values of 0.8, 0.1, and 0.01 are used to record the different types of bounding boxes. If the detecting confidence value is above 0.1, the box label is set to be 2. If it is between 0.1 and 0.01, the box label is 1. While the value is less than 0.01, the box label remains to be 0. According to the box label depicted above, hard samples of the cropped images is partitive.
\begin{figure}
\centering
\includegraphics[width=0.85\linewidth]{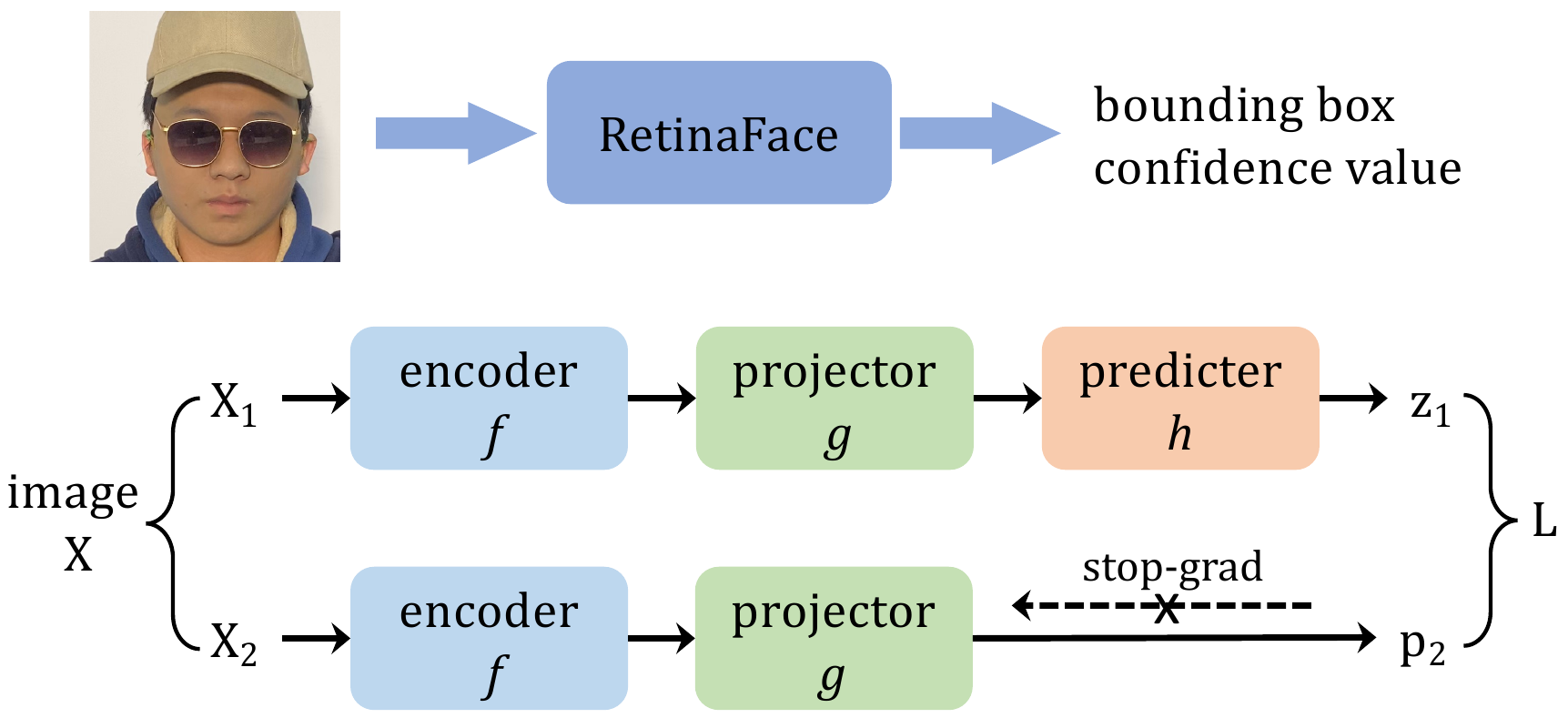}
\caption{\small{The framework of team inspire. Raw images are firstly processed in the upper flow. After that, this team used a Context Contrastive Learning framework to train, while the backbone is a SE-ResNeXt101 network.}}
\label{fig:inspire}
\end{figure}

For the training stage, SE-ResNeXt101 \cite{xie2017aggregated} was selected as the backbone. Besides, the team applied the Context Contrastive Learning(CCL) \cite{liu2021contrastive} architecture as the framework, which is shown in Fig.~\ref{fig:inspire}. As a result, they used a sampling strategy the same as that in~\cite{liu2021contrastive}. The MSE loss $L_{MSE}$, Cross Entropy loss $L_{CE}$ and Contrastive loss \cite{hadsell2006dimensionality} $L_{Contra}$ are applied to calculate total loss by the following weights:
\begin{equation}
\begin{aligned}
    L= L_{MSE} + L_{CE} + 0.7 * L_{Contra}
     \label{con:inspire-loss}
\end{aligned}
\end{equation}

Afterward, Ranger optimizer\footnote{\url{https://github.com/lessw2020/Ranger-Deep-Learning-Optimizer}} is set as a learning strategy with an initial learning rate of 0.001. The total epoch is 70, and the learning rate decays by 0.1 at 20, 30, 60 epochs, respectively.

\begin{figure}
\centering
\includegraphics[width=0.9\linewidth]{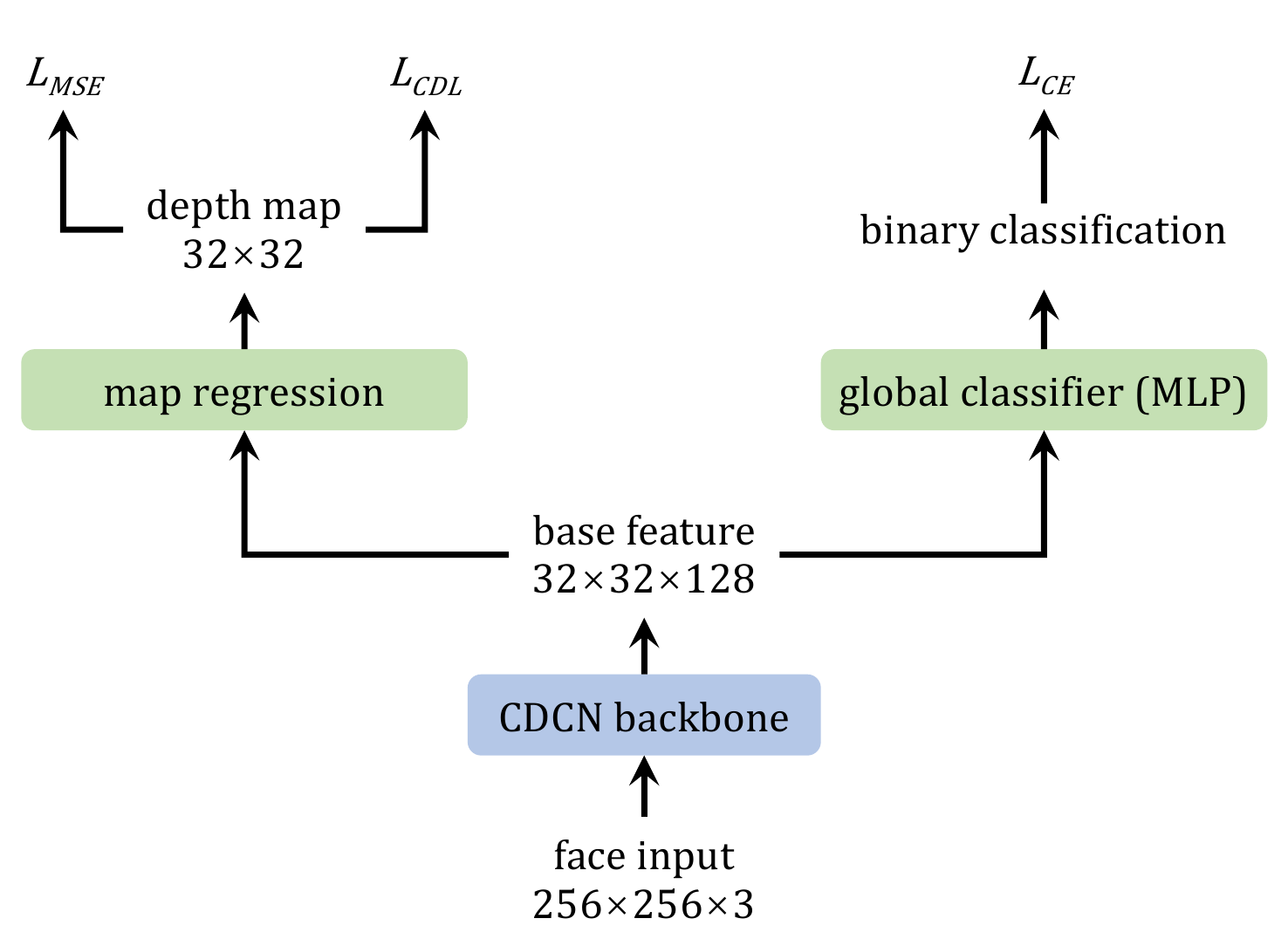}
\caption{\small{The framework of team Piercing Eye. Two branches are attached to the CDCN backbone, called map regression and global classifier, respectively.}}
\label{fig:Piercing-Eye}
\end{figure}

\vspace{0.5em}
\noindent \textbf{Piercing Eye} \quad
The team Piercing Eye used the modified CDCN~\cite{yu2020searching} as the basic framework, shown in Fig.~\ref{fig:Piercing-Eye}. During data processing, the face regions are detected from the original images, which are resized to $256\times 256$ and randomly cropped to $228\times 228$. Same as other teams, some types of data augmentation like color jitter are used.

\begin{figure}
\centering
\includegraphics[width=0.85\linewidth]{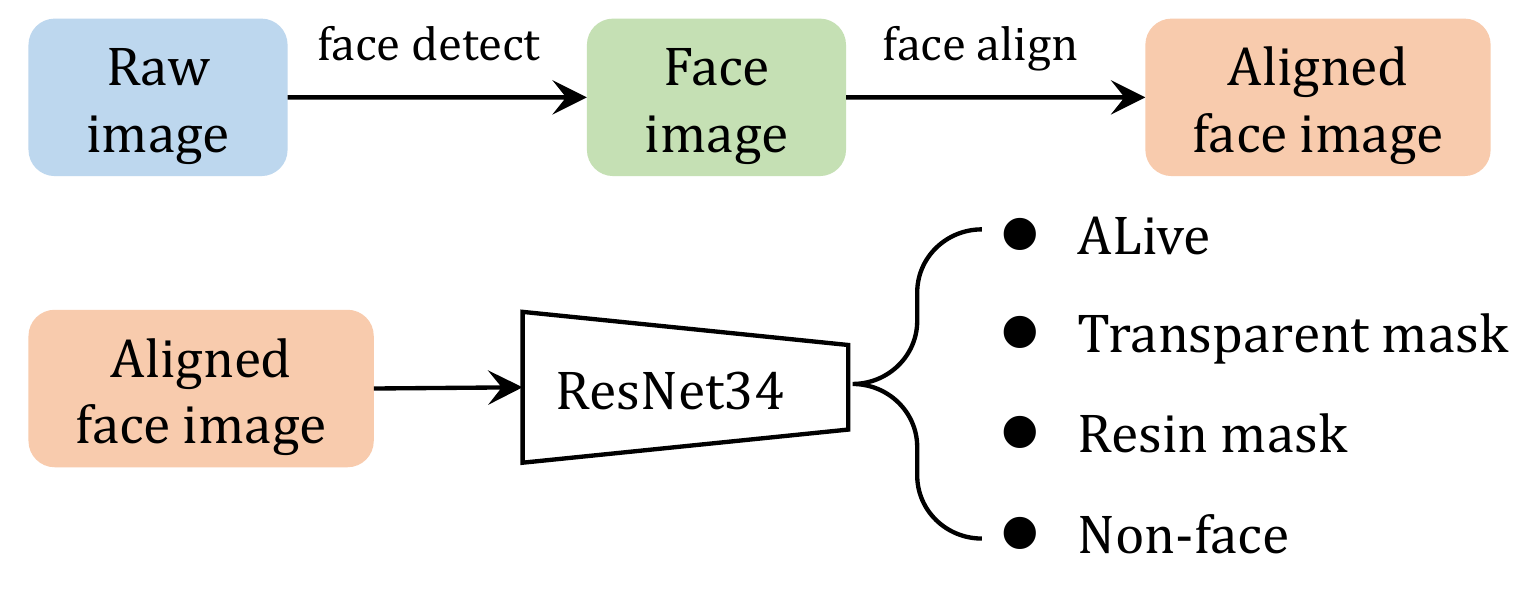}
\caption{\small{The framework of team msxf\_cvas. Raw images are firstly processed by detecting face and alignment. After that, a ResNet34 network is utilized to classify the input image into four types including live, transparent mask, resin mask and no face.}}
\label{fig:msxf}
\end{figure}

\begin{figure*}
\centering
\includegraphics[width=0.9\linewidth]{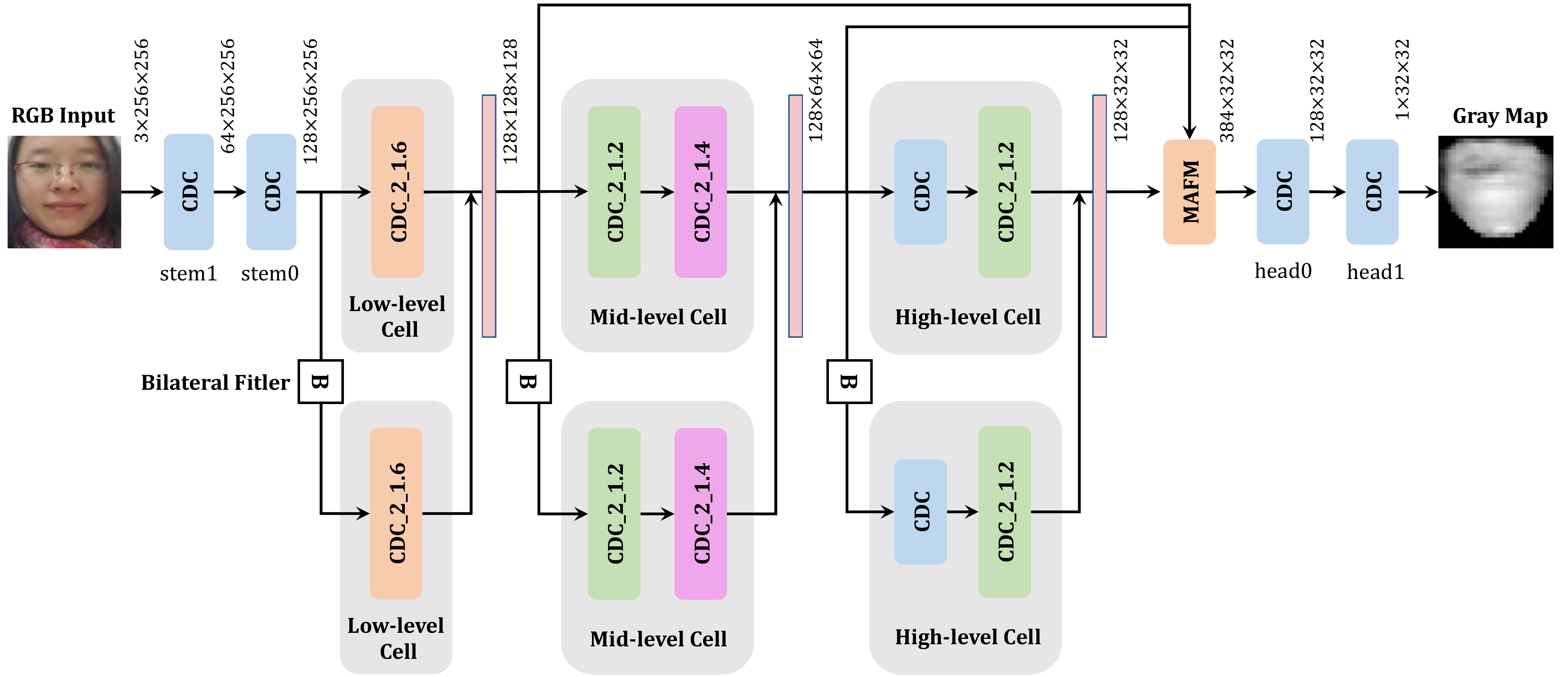}
\caption{\small{The framework of team VIC\_FACE. }}
\label{fig:VIC}
\end{figure*}

In addition to the original output (depth map) of CDCN, a multi-layer perceptron (MLP) is attached to the backbone, implementing the global binary classification. The shape of the depth map is $32\times 32$. The label of the real face region is set to 1, while the background and fake face region is set to 0. They trained the model with an SGD optimizer for 260 epochs using an initial learning rate of 0.002 and decreasing it by a factor of 0.5 with milestones. As in \cite{yu2020searching}, both mean square error loss $L_{MSE}$ and contrastive depth loss $L_{CDL}$ are utilized for pixel-wise supervision. They also perform cross-entropy loss in a global branch, denoted as $L_{CE}$. So the overall loss function is formulated as 
\begin{equation}
\begin{aligned}
    L= 0.5 * L_{MSE} + 0.5 * L_{CDL} + 0.8 * L_{CE}
     \label{con:Piercing-loss}
\end{aligned}
\end{equation}

\noindent \textbf{msxf\_cvas} \quad
From the analysis of competition data, the team finds two different distributions of spoof masks which are transparent material and fidelity material. They consider two materials (plaster and resin) as one category as the features of these two types looks similar. Besides, there are small amounts of noisy data without a human face which do not contain spoof or live features. Therefore, the team try to classify them as one category called non-face. The final task is to classify all data into four categories which are the live, transparent mask, resin mask, and non-face. Considering that there are many extreme posture and light and low-quality data in the competition data, they focus more on data augmentation strategies during training including cutMix, ISONoise, randomSunFlare, randomFog, motionBlur, and  imageCompression.

First of all, the team applied a face detector to detect faces and align faces by five points. After that, the mmclassification\footnote{\url{https://github.com/open-mmlab/mmclassification}} project was used to train a face anti-spoofing model. To begin with, the team chose a ResNet34 \cite{he2016deep} as the backbone and the cross-entropy loss was selected as the loss function. The whole framework is illustrated in Fig.~\ref{fig:msxf}.

\vspace{0.5em}
\noindent \textbf{VIC\_FACE}   \quad
The prerequisites need to know is that deep bilateral has been successfully applied in convolutional networks to filter the deep features instead of original images. Inspired by this, team VIC\_FACE proposed a novel method based on fusing the deep bilateral operator on the basis of original CDCN in order to learn more intrinsic features via aggregating multilevel bilateral macro- and micro- information. As shown in Fig.~\ref{fig:VIC}, the backbone model is an initial CDCN, which divides the backbone into multilevel (low-level, mid-level, and high-level) blocks to predict the gray-scale facial depth map with size $1\times 32\times 32$ from a single RGB facial image with size $3\times 256\times 256$. Besides, the DBO as a channel-wise deep bilateral filtering mimics a residual layer embedded in the network and replaces the original convolution layer by representing the aggregated bilateral base and residual features.

Specifically, at the first stage, they detect and crop the face area in the full image as input of the model. Secondly, they edit the images randomly with down-sampling and jpeg compression, which often occur unintentionally when the images are captured from different devices. Moreover, it is worth mentioning that excepting some regular data augmentation methods including cutout, color jitter and erase to improve generalization of the model, affine transformation of brightness and color of random area based on OpenCV is applied to simulate light condition in training data. Finally, they design a contrastive loss function for controlling the contrast depth map of gray-scale output and a mean-square error loss function for reducing the difference between augmented input and binary mask, then combining them into one loss function with Adam optimizer.

\begin{figure}
\centering
\setlength{\tabcolsep}{8pt}
\includegraphics[width=0.95\linewidth]{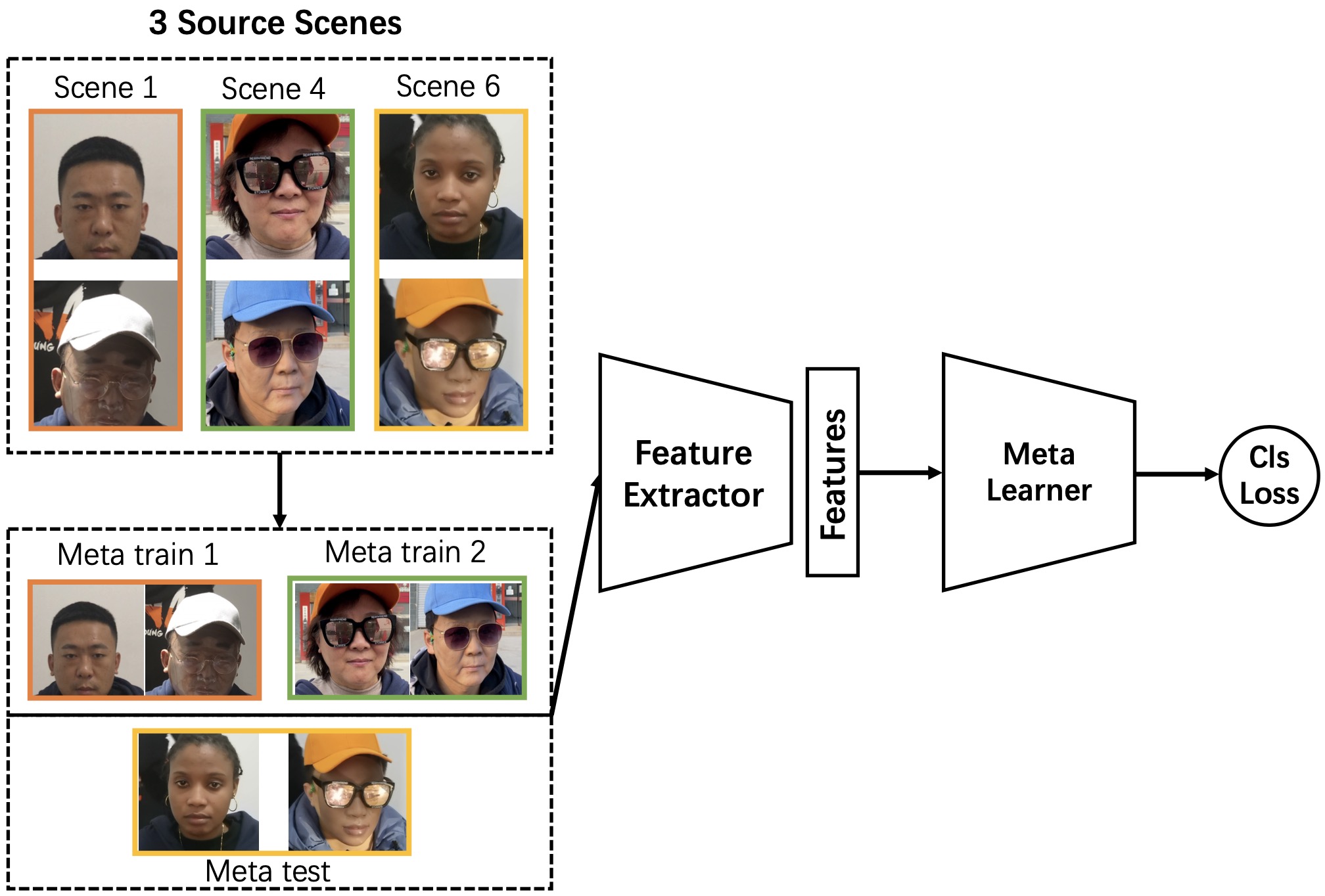}
\caption{The framework of DXM-DI-AI-CV-TEAM.}
\label{fig:DXM}
\vspace{-1.0em}
\end{figure}

\vspace{0.5em}
\noindent \textbf{DXM-DI-AI-CV-TEAM} \quad
Due to the generalization performance of the challenge evaluation algorithm for unknown attack scenarios, this team casts faces anti-spoofing as a domain generalization (DG) problem. To let the model generalize well to unseen scenes, inspired by~\cite{Shao_2020_AAAI}, the proposed framework trains their model to perform well in the simulated domain shift scenarios, which is achieved by finding generalized learning directions in the meta-learning process. Different from~\cite{Shao_2020_AAAI}, the team removed the branch of depth prior knowledge from the real face and mask, which contained similar depth information. Besides, a series of data augmentation and training strategies are used to achieve the best results.

In the challenge, the training data are collected in 3 scenes, namely White Light, Outdoor Sunshine, and Motion Blur (short for 1, 4, 6). Therefore, the objective of DG for this challenge is to make the model trained on the 3 source scenes can generalize well to unseen attacks from the target scene. To this end, as shown in Fig.~\ref{fig:DXM}, the framework in this team that composes of a feature extractor and a meta learner. At each training iteration, they divide the original 3 source scenes by randomly selecting 2 scenes as meta-train scenes and the remaining one as the meta-test scene. In each meta-train and meta-test scene, meta learner conducts the meta-learning in the feature space supervised by the image and label pairs denoted as $x$ and $y$, where $y$ are ground truth with binary class labels ($y=0/1$ is the label of fake/real face). In this way, their model can learn how to perform well in the scene shift scenarios through many training iterations and thus learn to generalize well to unseen attacks.

\section{Challenge Results}
\subsection{Challenge Results Report}
We adopted four metrics to evaluate the performance of the solutions, which are APCER, NPCER, ACER, and AUC respectively. Please note that although we report performance for a variety of evaluation measures, the leading metric was ACER. See from the Tab.~\ref{table:team-results}, which lists the results and ranking of the top 18 teams, we can draw three conclusions: (1) The ACER performance of the top 3 teams is relatively close, and the top 2 teams have the best results in all metrics. (2) The top 6 teams are from industry, which indicates that mask attack detection is no longer limited to academia, but also an urgent problem in practical application. (3) The ACER performance of all teams is evenly distributed between $3\%$ and $10\%$, which not only shows the rationality and selectivity of our challenge but also demonstrates the value of HiFiMask for further research.

\begin{table}[t]
\centering
\caption{Team and results are listed in the final ranking of this challenge.}
\scalebox{0.7}{
\begin{tabular}{|c|c|c|c|c|c|c|c|}
\hline  
R. & Team Name         & FP           & FN           & APCER     & BPCER      & ACER       & AUC  \\ \hline \hline
1       & VisionLabs                                                   
& 492          & \textbf{101} & 3.777          & \textbf{2.330} & \textbf{3.053} & \textbf{0.995} \\ \hline
2       & WeOnlyLookOnce                                               
& \textbf{242} & 193          & \textbf{1.858} & 4.452          & 3.155          & \textbf{0.995} \\ \hline
3       & CLFM                                                         
& 483          & 118          & 3.708          & 2.722          & 3.215          & 0.994          \\ \hline
4       & oldiron666                                                   
& 644          & 115          & 4.944          & 2.653          & 3.798          & 0.992          \\ \hline
5       & Reconova-AI-LAB                                              
& 277          & 276          & 2.126          & 6.367          & 4.247          & 0.991          \\ \hline
6       & inspire                                                      
& 760          & 176          & 5.834          & 4.060          & 4.947          & 0.986          \\ \hline
7       & Piercing Eyes                                       
& 887          & 143          & 6.809          & 3.299          & 5.054          & 0.983          \\ \hline
8       & msxf\_cvas                                                   
& 752          & 232          & 5.773          & 5.352          & 5.562          & 0.982          \\ \hline
9       & VIC\_FACE                                                    
& 1152         & 104          & 8.843          & 2.399          & 5.621          & 0.965          \\ \hline
10      & \begin{tabular}[c]{@{}c@{}}DXM-DI-\\ AI-CV-TEAM\end{tabular} 
& 1100         & 181          & 8.444          & 4.175          & 6.310          & 0.970          \\ \hline
11      & fscr                                                         
& 794          & 326          & 6.095          & 7.520          & 6.808          & 0.979          \\ \hline
12      & VIPAI                                                        
& 1038         & 268          & 7.968          & 6.182          & 7.075          & 0.976          \\ \hline
13      & reconova-ZJU                                                 
& 1330         & 183          & 10.210         & 4.221          & 7.216          & 0.974          \\ \hline
14      & sama\_cmb                                                    
& 1549         & 188          & 11.891         & 4.337          & 8.114          & 0.969          \\ \hline
15      & Super                                                        
& 780          & 454          & 5.988          & 10.473         & 8.230          & 0.979          \\ \hline
16      & ReadFace                                                     
& 1556         & 202          & 11.944         & 4.660          & 8.302          & 0.965          \\ \hline
17      & LsyL6                                                        
& 2031         & 138          & 15.591         & 3.183          & 9.387          & 0.951          \\ \hline
18      & HighC                                                        
& 1656         & 340          & 12.712         & 7.843          & 10.278         & 0.966          \\ \hline
\end{tabular}
}
\label{table:team-results}
\end{table}

\subsection{Competition summary and Future Work}
Through the introduction and result analysis of team methods in the challenge, we summarize the effective ideas for mask attack detection: (1) At the data level, data expansion is almost the strategy adopted by all teams. Therefore, data augmentation plays an important role in preventing the over-fitting of the model and improving the stability of the algorithm. (2) The segmentation of the face region can not only enlarge the local information to mine the difference between high fidelity mask and live face but also avoid the extraction of irrelevant features such as face ID. (3) Multi-branch-based feature learning is a framework widely used by participating teams. Firstly, a multi-branch network is used to mine the differences between the mask and live face from multiple aspects, such as texture, color contrast, material difference, etc., and then feature fusion is used to improve the robustness of the algorithm.

Since the challenge prohibits the use of additional datasets and pre-trained models, the performance is limited to a certain extent. In the following work, we further improve the performance from the following aspects: (1) Under visible light, it is difficult to distinguish between a live face and a mask. Therefore, we will use additional or generate multi-modal data~\cite{liu2021face,tang2021total} to assist mask attack detection. (2) Besides CNN, we will explore the effectiveness of recent vision transformer~\cite{dosovitskiy2021image} and MLP-like~\cite{tolstikhin2021mlp} architectures for face mask detection task. (3) As the HiFiMask dataset contains challenging dynamic lighting and scenes, we will explore more reliable rPPG~\cite{liu2021multi,yu2021transrppg} technology for detecting liveness clues.

\section{Conclusion}
We organized the \emph{3D High-Fidelity Mask Face Presentation Attack Detection Challenge at ICCV2021} based on the HiFiMask dataset and running on the CodaLab platform. 195 teams registered for the competition and 18 teams made it to the final stage. Among the latter, teams were formed by 12 companies and 6 academic institutes/universities. We first described the associated dataset, the challenge protocol, and the evaluation metrics. Then, we reviewed the top-ranked solutions and reported the results from the final phases. Finally, we summarized the relevant conclusions, and pointed out the effective methods against mask attacks explored by this challenge.

\section{Acknowledgement}
This work was supported by the Chinese National Natural Science Foundation Projects $\#$61961160704, $\#$61876179, the External cooperation key project of Chinese Academy Sciences $\#$ 173211KYSB20200002, the Key Project of the General Logistics Department Grant No.AWS17J001, Science and Technology Development Fund of Macau (No.~0010/2019/AFJ, 0008/2019/A1, 0025/2019/AKP, 0019/2018/ASC), by the Spanish project PID2019-105093GB-I00 (MINECO/FEDER, UE), and by ICREA under the ICREA Academia programme.

{\small
\bibliographystyle{ieee_fullname}
\bibliography{egbib}
}

\end{document}